\documentclass[runningheads]{llncs}
\usepackage{multirow}
\usepackage{graphicx}
\usepackage{amsmath}
\usepackage{xcolor}
\usepackage{caption}
\usepackage{subcaption}
\usepackage[misc]{ifsym} 
\begin{document}
%
%\title{Domain Adaptation for Cardiovascular Disease Risk Estimation From Retinal Fundus Images}
\title{Camera Adaptation for Fundus-Image-Based CVD Risk Estimation}
%\title{Contribution Title\thanks{Supported by organization x.}}
\author{Zhihong Lin\inst{1}\and
% index{Lin, Zhihong} 
Danli Shi\inst{2}\and
% index{Shi, Danli} 
Donghao Zhang\inst{1}\and
% index{Zhang, Donghao} 
Xianwen Shang\inst{3}\and
% index{Shang, Xianwen} 
Mingguang He\inst{3}\and
% index{He, Mingguang} 
Zongyuan Ge\inst{1}$^{(\textrm{\Letter})}$
% index{Ge, Zongyuan} 
}
\authorrunning{F. Author et al.}
\institute{Monash University, Clayton, VIC, Australia
\email{zongyuan.Ge@monash.edu}\\
\and
Zhongshan Ophthalmic Center, Sun Yat-sen University, Guangzhou, China\\
\and
University of Melbourne, Parkville, VIC, Australia\\
}
\maketitle              % typeset the header of the contribution
%Studies show that deep learning has the potential to estimate CVD risk from the fundus photograph to support early screening. 
\begin{abstract}
%The application will also significantly reduce the cost of CVD risk screening and improve healthcare democratization. 
%With the conventional deep learning methodology, the prediction risk score of one patient can be variant on different imaging devices. In a clinical scenario, the inconsistency can be harmful to user acceptance. 
%imaging settings between the research-oriented proof of concept (POC) and. such as Topcon and quality
%The application will also significantly reduce the cost of CVD risk screening and improve healthcare democratization. 
%With the conventional deep learning methodology, the prediction risk score of one patient can be variant on different imaging devices. However, in a clinical scenario, the inconsistency can harm user acceptance. 
Recent studies have validated the association between cardiovascular disease (CVD) risk and retinal fundus images. Combining deep learning (DL) and portable fundus cameras will enable CVD risk estimation in various scenarios and improve healthcare democratization. However, there are still significant issues to be solved. One of the top priority issues is the different camera differences between the databases for research material and the samples in the production environment. Most high-quality retinography databases ready for research are collected from high-end fundus cameras, and there is a significant domain discrepancy between different cameras. To fully explore the domain discrepancy issue, we first collect a Fundus Camera Paired (FCP) dataset containing pair-wise fundus images captured by the high-end Topcon retinal camera and the low-end Mediwork portable fundus camera of the same patients. Then, we propose a cross-laterality feature alignment pre-training scheme and a self-attention camera adaptor module to improve the model robustness. The cross-laterality feature alignment training encourages the model to learn common knowledge from the same patient's left and right fundus images and improve model generalization. Meanwhile, the device adaptation module learns feature transformation from the target domain to the source domain. We conduct comprehensive experiments on both the UK Biobank database and our FCP data. The experimental results show that the CVD risk regression accuracy and the result consistency over two cameras are improved with our proposed method. The code is available here: \url{https://github.com/linzhlalala/CVD-risk-based-on-retinal-fundus-images}
%We will release our code to promote research in this field. 

% Moreover, our camera adaptation module achieves the low-cost adaptation target with only a small amount of extra data. 
%enhance the model performance and consistency. 
%With the cross-laterality feature alignment training scheme, we let the model focus on learning the common knowledge in the left and right fundus images of the same patient. 
% low-cost adaption: camera, less data,  medical democratization

\end{abstract}
\section{Introduction}\label{intro}
% QRISK3, 2016 European Guidelines, 

Cardiovascular diseases (CVD) are the leading causes of mortality around the world~\cite{roth2020global}. Early screening and identification of the at-risk populations are necessary for CVD prevention and control. The existing CVD risk estimation guidelines (for example, the WHO-CVD score~\cite{kaptoge2019world}) are based on the patient physical measures and laboratory testing, such as body massive index~(BMI) and high-density lipoprotein~(HDL). However, some inputs are difficult to acquire in scenarios where medical professionals and laboratory testing are not accessible. 
%in outreach programs, mobile eye care, and satellite offices, 
%The microvascular system captured by the retinal images can also show the feature of CVD initiation and progression. 
A prospective solution is to predict fundus-image-based CVD risk with an automated CAD algorithm using a portable fundus camera. Recent studies~\cite{cheung2012retinal,poplin2018prediction} have revealed the associations between the retinal fundus and its CVD risk factors. 
%Furthermore, the deep learning technology has been validated to predict CVD risk factors from the fundus images~\cite{poplin2018prediction}.
%On the other hand, with the advantage of simple operation and low cost, the portable fundus camera is ideal for outreach programs and mobile CVD risk screening. Hence, 
The fundus-image-based CVD risk estimation on a portable camera can significantly reduce the examination cost and the requirement of medical professionals. It will also improve medical democratization and let more people benefit from medical AI. 
% \ZY{All the contents are here, however the logic and conjunction upon this point are not correct, please edit these sentences again and improve the flow.}
%~\footnote{\underline{https://www.ukbiobank.ac.uk/}}
Despite this prospective application, there still exist issues to be solved to achieve successful implementation. One of the most challenging problems is the domain discrepancy between fundus cameras. The ideal high-quality retinal samples for CVD risk inference are mainly captured by expensive clinical devices. For example, the retinal images from the UK Biobank~\cite{sudlow2015uk}~(UKB) are collected using the Topcon 3D OCT-1000 MKII. Meanwhile, the retinal samples in the practical scenario are sometimes captured by low-cost portable fundus cameras. The key components of the fundus camera, such as light source, lens, and camera sensor, are all different. The impacts on images quality and style can lead to the problem of domain discrepancy. 
% \ZY{Please give 1-2 sentences on camera hardware spec, lens, image quality differences. You can ask Danli and Medway for more information.}
% Therefore, to overcome the domain shift is a key essential to develop adapt train the model from a large database to the hand-held camera is. 
% within a small amount of new collection data. 
%The UK Biobank dataset contains physical measures, laboratory testing, and long-term tracking of disease events. It is an ideal training material for the CVD risk estimation from the fundus image.
%According to our performance comparison, we select the vision transformer (ViT) model~\cite{augreg} as the baseline model. At the same time, we find there is also inconsistency between the images of the same patient and the same camera but different eyes. The inconsistency can harm the acceptance from the end-user in the future application stage. 
%As the two cameras can be regarded as two domains and the target domain label is unavailable, the issue can be defined as an unsupervised domain adaptation problem.

To tackle the problem of fundus camera adaptation, the existing efforts have mainly focused on feature and image alignment. The CFEA~\cite{liu2019cfea} exploited collaborative adversarial learning and self-ensembling for feature adaptation on the optic disc~(OD) and optic cup~(OC) segmentation task. The CSFA~\cite{opticseg2022lei} concentrated on the content and style feature consistency among source domain images, target-like query images and target domain images also for the disc and cup segmentation task. Yang \textit{et al.}~\cite{yang2020residual} proposed a camera-oriented residual-CycleGAN to pre-process the images and improve the diabetic retinopathy prediction performance on multiple cameras data. Ju \textit{et al.}~\cite{uwf2021julie} proposed a modified CycleGAN to bridge the domain gap between regular and ultra-widefield fundus images for several tasks. However, these solutions only consider the target domain performance with abundant training images from the target domain required, which is impossible to collect in our task. Furthermore, these methods also ignore the cross-domain consistency, which may lead to disastrous outcomes for CVD risk evaluation.
%However, our task tends to be evaluated by the cross-domain consistency because of the difficulty to obtain CVD risk ground truth. 
%These solutions are evaluated by the target domain performance.
%TODO,  \ZY{I do not see domain ADA is a limitation, please give more reasonable points for the incompleteness of those methods or dataset setting}
%We hypothesis that domain generalization can also benefit camera adaptation by providing a robust and adaptive source model. \ZY{With little information provided here, I am confused between the domain adaptation vs. generalization? }
%of the previous efforts in unsupervised domain adaptation can be divided into two categories: The feature-alignment-based methods such as DANN~\cite{ganin2015unsupervised} encourage the model to learn domain invariant feature representation between domains. The image-alignment-based methods use the GAN~\cite{goodfellow2014generative}-based method to translate the image from the target domain to the source domain. There are also feature learning method~\cite{ahn2020unsupervised} applying extra networks to transform the features. However, these methods (TODO, the reference are not detailed enough, make the limit not accurate, move more to a camera problem, will make it easier?).
%are overall focusing on the feature distribution but neglecting the model generalization.
%on the fundus-based CVD risk, robust and adaptive,  with a minimum cost

%To establish the benchmark and observe the problem, 
In this work, we collect a Fundus Camera Paired~(FCP) dataset containing the pair-wise fundus images of the same patients using two cameras~(Topcon upright camera and Mediwork portable camera). 
\footnote{We do not provide CVD risk ground-truth for the FCP dataset due to the absence of patient information.}
% we investigate the domain discrepancy problem in a task-oriented plan. Our goal is to pre-train a model on the WHO-CVD score with the UKB bio-bank and then adapt the model to our target camera. 
Our strategy on the task is two-fold: domain generalization~\cite{wang2021generalizing} and domain adaptation~\cite{medicalda}. The domain generalization is to improve the adaptability with the unseen target domain. Inspired by the contrastive learning~\cite{SimSiam,chen2021mocov3,BYOL}, we propose a cross-laterality feature alignment~(CLFA) pre-training scheme to utilize the images of both lateralities, which are believed to share the invariant representation for the CVD risk.
%\ZY{Give a short description of why it works for your task 
Besides the supervised CVD risk regression, we introduce an asymmetric feature alignment task for domain generalization by comparing the pair-wise image features and letting the superior one teach the other one to enhance the learning on domain-invariant representation.
Domain adaptation is to adapt the model to the target domain given. To maintain the model knowledge on the CVD task, we adopt a plug-in camera adaptor module based on multi-head self-attention. The camera adaptor can transform the target domain image feature to match its source domain edition. 
The experiment shows that our two-step strategy improves the CVD risk regression performance on the UKB dataset and the prediction consistency between the two cameras on our FCP dataset. Moreover, we conduct ablation studies on our pre-training method and camera adaptor. We find that our CLFA pre-training reduces the feature-space discrepancy between the UKB dataset and the other cameras. Meanwhile, our camera adaptor can significantly improve the CVD result coefficient on different pre-trained models. 
\section{Method}
% As introduced above, our proposed method has two steps. The first step is the backbone training with the labeled CVD risk dataset. The second step is training the device adaptation module with our Fundus Camera Paired (FCP) dataset while the backbone ViT model is frozen. The module is trained to align the image feature from the hand-held camera to the image feature from the Topcon camera. 
%Our method has two parts: the pre-training scheme to enhance the backbone model generalization and the camera adaptor module to perform adaptation.

\subsection{Cross-Laterality Feature Alignment Pre-training}\label{sec_training}
\begin{figure}[b]
\includegraphics[width=\textwidth]{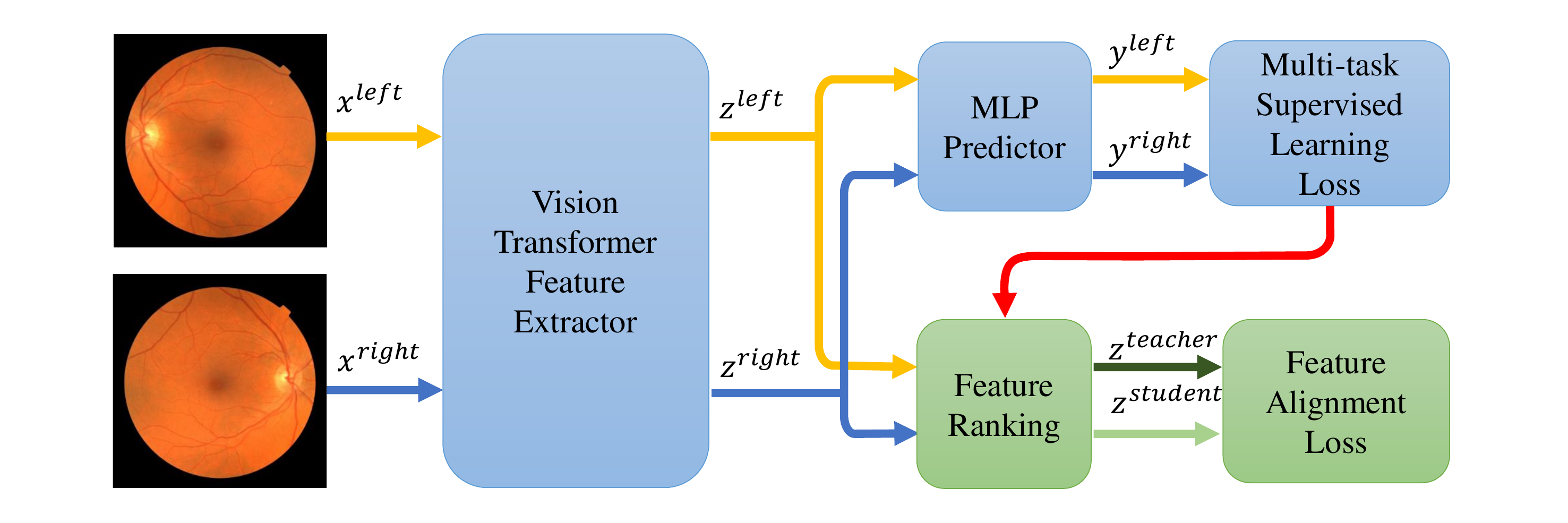}
\caption{The cross-laterality feature alignment pre-training has two branches: a multi-task supervised learning branch and an asymmetric feature alignment branch.}% \ZY{Expand on what each branch is; I think Multi-task regression is better than your current name.}} 
\label{coretraining}
\end{figure}

Our cross-laterality feature alignment~(CLFA) pre-training is inspired by the domain generalization research~\cite{wang2021generalizing} and contrastive learning~\cite{BYOL,SimSiam,chen2021mocov3}. The UKB provides both left and right fundus photos for each patient, which can be utilized as natural positive sample pairs. We hypothesize that the visual clues of CVD risk have invariant representation over the two eyes. A deep learning model would have better generalization when identifying the invariant representation. As shown in Fig.\ref{coretraining}, our pre-training adopts the siamese network~\cite{chicco2021siamese} and lets both lateralities share the backbone model (a vision transformer model in this paper). The model is simultaneously trained with two branches. 
%\ZY{The method is fine, but why only one CVD score is presented? I think you should change it to more like a self-supervised/mask learning, where you have CVD scores for both eyes; however, you intentionally mask one out, then try to train and align them.} 
%The deep learning model is expected to find out  \ZY{Find out is oral English, please change it to written English}. 

In the supervised learning branch, we jointly train the model on the WHO-CVD score as well as seven clinical variables explicitly related to the WHO-CVD, including age, systolic blood pressure~(SBP), total cholesterol~(TC), body massive index~(BMI), gender, smoking status, and diabetes status. The aim is to improve the model generalization by sharing representations between the tasks~\cite{zhang2021survey}. 
%Besides, we select seven c. Hence we have five regression targets (WHO-CVD score) and three binary classification labels (). 
%\ZY{I do not like such called tasks, they are very confusing. Please improve these sentences. Moreover, the motivation of this architecture or loss design is missing. Why did you design this way? How does this link to your overall task for adaptation?}ZY {x variables not denoted.}
Let $S = \left\{x_{i}^{left},x_{i}^{right},y_{i}^{rgs},y_{i}^{cls}\right\}_{i=1}^{n}(rgs=1,2,3,4,5; cls=1,2,3)$ denote the training set, where the $x_{i}^{left}$ denotes the left fundus photo, $x_{i}^{right}$ denotes the right fundus photo, $y_{i}^{rgs}$ denotes the regression labels, and $y_{i}^{cls}$ denotes the binary classification labels. The backbone model is defined as the function of $f_\theta:\mathcal{X}\rightarrow(\mathcal{Y}^{rgs}, \mathcal{Y}^{cls})$. 
%where the $\mathcal{Y}^{r}=\left\{y_{i}^{r}\right\}_{i=1}^{n}(r=1,2,...,R)$ denotes the regression targets and the $\mathcal{Y}^{c}=\left\{ \right\}_{i=1}^{n}(c=1,2,...,C)$ denotes the binary classification labels.
The mean squared error~(MSE) and binary cross-entropy~(BCE) are used as the loss functions for the regression and binary classification. The weighting of tasks are denoted as $W=\left\{w^{rgs}, w^{cls}\right\}(r=1,2,...,5; c=1,2,...,3)$. The $\sigma(\cdot)$ denotes the $sigmoid$ function. The loss function of branch is defined as follows:
\begin{equation}\label{eq1}
\begin{split}
    % L_{S}^{sup.}(\theta)= &\frac{1}{n}\sum_{i=1}^{n}\Bigg\{\sum_{r=1}^{R}w^{r}\Big(y_{i}^{r}-f_\theta^{r}(x_{i})\Big)^{2}\\
    % &-\sum_{c=1}^{C}w^{c}\bigg[\Big(y_{i}^{c}\cdot log\sigma (f_\theta^{c}(x_{i}))+(1-y_{i}^{c})\cdot log\big(1-\sigma (f_\theta^{c}(x_{i}))\big)\Big)\bigg]\Bigg\}
    %\frac{1}{n}\sum_{i=1}^{n}\Bigg\{\Bigg\}
    \ell_{sup.}(\theta)=\sum_{rgs=1}^{5}w^{rgs}\cdot MSE\Big(y_{i}^{rgs}, f_\theta^{rgs}(x_{i})\Big) +\sum_{cls=1}^{3}w^{cls}\cdot BCE\Big(y_{i}^{cls},  \sigma(f_\theta^{cls}(x_{i}))\Big)
\end{split}
\end{equation}
Note that both eyes are used as laterality-agnostic inputs. Hence, for every patient we have two supervised loss $\{\ell_{sup.}^{left},\ell_{sup.}^{right}\}$.

The feature alignment learning branch is designed to enable the interaction between two fundus image features from the same patient. With the input pair $\{x_{i}^{left},x_{i}^{right}\}$, the two supervised learning loss $\{\ell_{sup.}^{left},\ell_{sup.}^{right}\}$ will have a larger one and a smaller one. 
The feature with the smaller loss is selected as the teacher while the other is selected as the student. This operation can be regarded as a feature-level knowledge distillation. The backbone model function can be split into two steps: the feature extractor function $g_\theta:\mathcal{X}\rightarrow\mathcal{Z}$ and the predict function $p_\theta:\mathcal{Z}\rightarrow(\mathcal{Y}^{rgs}, \mathcal{Y}^{cls})$. 
The $SD(\cdot)$ means the stop-gradient operation. The loss function is defined as follows:
%ZY{This paragraph is clear, but should appear at the begining of this section.} 
\begin{equation}\label{eq2}
%\begin{split}
\ell_{ali.}(\theta)=
\begin{cases}
MSE\big(g_\theta(x_i^left), SD(g_\theta(x_i^riht))\big) & \ell_{sup.}^{left}\ge\ell_{sup.}^{right},\\
MSE\big(g_\theta(x_i^right), SD(g_\theta(x_i^left))\big) & \ell_{sup.}^{left}<\ell_{sup.}^{right}
\end{cases}
%\end{split}
\end{equation}
% The contrastive learning is based on a pretext task of discriminating the positive and negative sample pairs. The recent self-supervised learning research BYOL~\cite{BYOL}, the SimSiam~\cite{SimSiam} has verified that the contrastive learning can be done with only the positive pairs. The MoCo v3~\cite{chen2021mocov3} has verified that the ViT can also be the backbone encoder in self-supervised learning. However, their work are all using two image augment editions as a positive pair. Due to our task and data nature, we find another approach that we can treat the patient as the sampling objective and use the fundus images from the two eyes as a positive pair. Similar to the SimSiam, we have a backbone model and an extra multilayer perceptron (MLP) prediction head. The backbone model function can be split into two steps: . The MLP head is denoted as $h_\theta: \mathcal{Z}\rightarrow\mathcal{P}$ and the $SD(\cdot)$ means the stop-gradient operation. Let $S = \left\{ x_i^l,x_i^r\right\}_{i=1}^{n}$ denote the traing set and the $\mathcal{D}$ denotes the cosine similarity loss. The loss function of the contrastive learning branch is defined as follows:
% \begin{equation}\label{eqx}
% \begin{split}
%     L_{S}^{con.}(\theta)= &\frac{1}{n}\sum_{i=1}^{n}\Bigg\{\frac{1}{2}\mathcal{D}\Big(SD(g_\theta(x_i^l)),h_\theta(g_\theta(x_i^r))\Big)+\frac{1}{2}\mathcal{D}\Big(SD(g_\theta(x_i^r)),h_\theta(g_\theta(x_i^l))\Big)\Bigg\}
% \end{split}
% \end{equation}
Hence the overall loss function of the CLFA pre-training is as follows, where the $\lambda$ denotes the weighting for the feature alignment branch:
\begin{equation}\label{eq3}
    L_{S}(\theta)=L_{S}^{sup.}(\theta) + \lambda\cdot L_{S}^{ali.}(\theta)
\end{equation}
\subsection{Self-Attention Camera Adaptor Module}\label{adaption}
Our self-attention camera adaptor~(SACA) module is inspired by the feature learning approaches~\cite{medicalda} in domain adaptation research. We freeze the pre-trained model to anchor the outcome of the source domain. Meanwhile, we add an adaptor module for the target domain. This strategy allows us to utilize the pairing information in our FCP dataset and establish the benchmark based on cross-domain consistency. 
%The model will lose its core function in the source domain and make the whole task unreliable.
%and the GAN-based methods' generation images show serious information loss. 
%In our early investigation, we tested several unsupervised domain adaptation methods. We found the distribution-discrepancy-based methods such as DAN~\cite{long2015learning} have a low prediction consistency over pair-wise data despite they do reduce the overall distribution discrepancy. Hence,

\begin{figure}
\includegraphics[width=\textwidth]{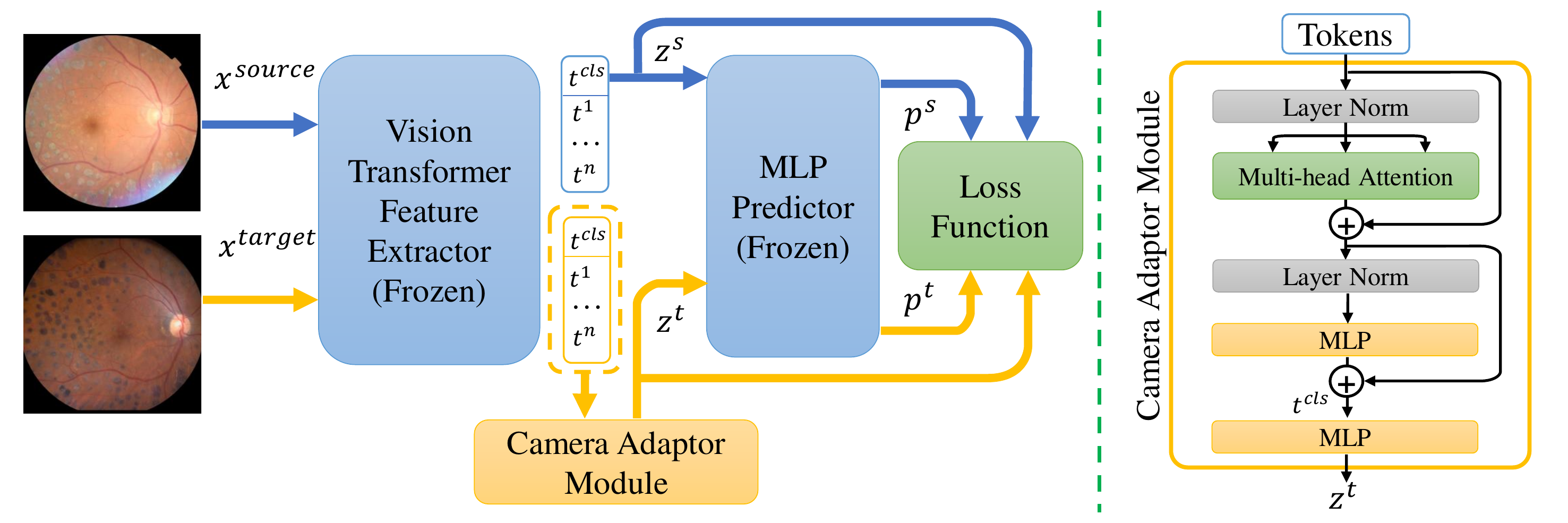}
\caption{The left part of the figure illustrates the workflow of prediction. The blue arrows represent the prediction procedure of the source domain, while the yellow arrows represent the prediction procedure of the target domain. The right part of the figure shows the structure of the camera adaptor module.} 
\label{deviceadapation}
\end{figure}
%patient: guansongwen
The module network is shown in Fig.\ref{deviceadapation}. Based on our early performance comparison, we choose the vision transformer~\cite{dosovitskiy2021image} (ViT) as our backbone model. The ViT feature extractor outputs the classification token and image patch tokens as $T^{out}=\left\{t^{cls}, t^1, t^2,...,t^N\right\}$.
% \ZY{You suddenly jump to Transformer, however, not enough background information is given here.} 
% \begin{equation}\label{eq4}
% \begin{split}
% \end{split}
% \end{equation}
Generally, the classification token ($t^{cls}$) is used as the image feature (denoted as $\mathcal{Z}$ in the Sec.~\ref{sec_training}). However, the image patch embeddings are also informative and can be utilized in our task. We adopt the block design of ViT, including the multiheaded self-attention~(MSA), multilayer perceptron~(MLP), Layernorm~(LN), and residual connections. Besides the block, we apply an extra MLP projector. 
The camera adaptor module ($k_\beta:T\rightarrow\mathcal{Z}_{a}$) can be described as follow:
%when we use the adapted classification token as an adapted image feature. 
\begin{equation}\label{eq5}
\begin{split}
    &{T}'=MSA(LN(T_{out}))+T_{out}\\
    &T_{a}=MLP(LN({T}'))+{T}'=\left\{t_{a}^{cls}, t_{a}^1, t_{a}^2,...,t_{a}^N\right\}\\
    &\mathcal{Z}_{a} = MLP(t_{a}^{cls})
\end{split}
\end{equation}
%and it has two steps $q_\theta:\mathcal{X}\rightarrow T, h_\theta: t^{cls}\rightarrow\mathcal{P}$. Our device adaptation module can be plugged in as $q_\theta:\mathcal{X}\rightarrow T, q_\beta:T\rightarrow\mathcal{Z}_{DA},  h_\theta: t_{DA}^{cls}\rightarrow\mathcal{P}$.
%\ZY{See comment}
%\ZY{这段和上一段在一个section里，但是内容很割裂。}
%The camera adaptor module can be plugged into the source model without any impact to the source model parameters and 
To train the camera adaptor module, we use pair-wise data which is denoted as $P=\left\{x_{i}^{s},x_{i}^{t}\right\}_{i=1}^{n}$. For each image pair $\left\{x^{s},x^{t}\right\}$ fed in the source model and the adapted model, we have the WHO-CVD prediction $\left\{p^{s},p^{t}\right\}$. We define the training loss function as:
% The image feature distance and the result consistency are both considered. 
%outcome tokens $\left\{T^{s},T^{t}\right\}$, the features $\left\{z^{s},z^{t}\right\}$, and the
%are expected to be similar
\begin{equation}\label{eq6}
\begin{split}
    L_P(\beta)= \frac{1}{n}\sum_{i=1}^{N}MSE\Big(p_i^{s},p_i^{t}\Big)
    %MSE\Big((z_{i}^{s}), k_{\beta}(T_i^{t})\Big)
\end{split}
\end{equation}
%Note that the $\theta$ is frozen in adaptation training. In the implementation, we alternatively cache the $T_{out}$ for efficiency.
\section{Experiment}
\subsection{Dataset}
In this work, we use the UK Biobank retinal photography dataset~(UKB)~\cite{sudlow2015uk} for backbone model pre-training and our Fundus Camera Paired (FCP) dataset for camera adaptor training and evaluation.
%\ZY{Yes, having a name for your paired dataset is good, it should appear earlier in this article.}. 
The UKB originally had 58,700 patients, and the images were captured by a Topcon 3D OCT-1000 MKII camera. We conduct an image quality assessment and a clinical information assessment. 
%whose information is insufficient for WHO-CVD score calculation. 
The selected 41,530 patients are split into 33,224 for training and 8,306 for validation balanced by the WHO-CVD score. From the validate split, we slice a balanced subset~(UKB*) of 416 patients (741 images) for the feature distribution study.
%We conduct for the UKB dataset following Fu method~\cite{fu2019evaluation} and filter out the patient whose information (age, gender, SBP, total cholesterol, BMI, smoking status, and diabetes status) is incomplete to calculate the WHO-CVD score. 
% We calculate the WHO-CVD scores according to \cite{kaptoge2019world} and transform it into a logarithmic scale to normalize the distribution. The images are center-cropped and resized to $512\times512$. The pre-processed data is split into 33,224 for training and 8,306 for validation by the patient unit and balanced by the WHO-CVD score. 
The FCP dataset has 227 patients and 415 pairs of pair-wise photos captured by Topcon TRC-NW8 camera and the Mediwork FC-162 portable camera. The patients are split into 182 for training and 45 for validation randomly. There is no patient information collected in the FCP dataset. 
%d data is split into five folds randomly for cross-validation. 
%\ZY{Please present information in the following order, Data description and partition, evaluation metric. Compared method Fu should not be here. The imaging size should not be here; it should be in the implementation section. }
\subsection{Implementation Details and Metrics}
For the UKB dataset, we calculate the WHO-CVD score according to the WHO guideline~\cite{kaptoge2019world} and transform it into the logarithmic scale to normalize the distribution. 
Our backbone vision transformer (ViT) model structure is as the ``R26+S/32'' in the AugReg~\cite{augreg}. The model loads the ImageNet pre-trained~\cite{augreg} as the initial state. The default task weighting $w^c, w^r, \lambda$ mentioned in Sec.\ref{sec_training} are all set to 1.0 for default. 
For backbone model pre-training, the batch size is 16, and the start learning rate is 1e-4. The augmentation includes resize, crop, color jitter, and grayscale. 
For the camera adaptor module, the batch size is 32, and the start learning rate is 1e-2. 
All experiments in this study use the Adam optimizer with momentum 0.9 and weight decay 1e-4.
%(Hardware: NVIDIA GeForce 3090 GPU; Software: PyTorch 1.8.0, and Python 3.8.8). 
%implementation is adopted from the \texttt{timm} repository ~\cite{rw2019timm} and the. 
%The augmentation in backbone model pre-training includes resize, crop, color jitter, and grayscale.
%The input image size is $384\times384$. 
%The images are center-cropped and resized to $512\times512$.
%For the patients with only one photo available, we use the horizontally flipped edition to comprise a pair. 
%start learning rate 1e-4, and the batch size is 16. The experimental training is performed for 50 epochs. For the camera adaptation experiment, the training is via an Adam optimizer with a start learning rate of 1e-2, momentum 0.9, and weight decay 1e-4. The batch size is 32. The experimental training is performed for 200 epochs.
%In the following experiments, w

\textbf{coefficient of determination ($R^2$)} is used to evaluate the WHO-CVD regression performance and the WHO-CVD result consistency between cameras with the Topcon result as pseudo target and the Mediwork result as the prediction. We also experimentally introduce the \textbf{Multi-Kernel Maximum Mean Discrepancy (MK-MMD)~\cite{gretton2012optimal}} to measure the feature-distribution discrepancy between domains. 

\subsection{Ablation Study}
We perform an ablation study to show the effectiveness of each design in our backbone model pre-training and camera adaptation. 
\subsubsection{Backbone Model Pre-training}\label{ablation1}
In this experiment, we pre-train several backbone models on the UKB dataset and then train the camera adaptor~\footnote{As our uniform self-attention camera adaptor} for each backbone model. Tab.~\ref{ablatable1} shows our experiment result of the backbone model pre-training. The baseline model is the ViT model with only the supervised learning branch and the WHO-CVD regression task. In ``Weight1'' the eight tasks describe in Sec.~\ref{sec_training} are equally weighted. With ``Weight1'', we observe the $R^2_{CVD}$ degrades and the $R^2$ scores of systolic blood pressure~(SBP), total cholesterol~(TC), and body massive index~(BMI) regression are very low. Therefore, in ``Weight2,'' we remove the SBP, TC, BMI and increase the weight of WHO-CVD and achieve an increased $R^2_{CVD}$. Based on the ``Weight2'' supervised learning branch, we add the second branch of contrastive learning~\cite{SimSiam} or our Cross-Laterality Feature Alignment (CLFA). Surprisingly, the ``Weight2+SimSiam'' has obtained a collapsing performance while the ``Weight2+CLFA'' achieves the highest $R^2_{CVD}$ and post-adaptation $R^2_{(T,M)}$. It indicates that our ``Weight2+CLFA'' provides the best potential and adaptability. 
The MK-MMD comparison shows the ``Weight2+CLFA'' has the lowest feature-distribution discrepancy between UKB and Mediwork as well as between UKB and Topcon. These two variables show a negative correlation to the $R^2_{(T,M)}$. However, the pre-adaptation $R^2_{(T,M)}$ and feature-distribution discrepancy between Topcon and Mediwork shows no strong relation to post-adaptation $R^2_{(T,M)}$. The relation between MK-MMD and post-adaptation $R^2_{(T,M)}$ indicates the MK-MMD can be a metric to evaluate model adaptability in future research. 
%\ZY{what do you mean regulation between?}
% because their error is too high. \ZY{very confusing without variables references, do not show actual value, for w1 say equally weighted, then weight2 just say what var has been increased for weighting and what has been down-weighted.} % \noindent\textbf{Pre-training} To compare the standard contrastive learning method with our cross-laterality Feature Alignment (CLFA), we implement the SimSiam~\cite{SimSiam} for the secondary branch. 
%\ZY{This paragraph had lots of writing issues, please re-write.}
%and different feature alignment designs such as aligning to the mean feature (AMF) 

\begin{table}[t!]
\centering
\caption{Ablation study on the pre-training (CVD=WHO-CVD Regression
on the UKB; T=Topcon data in FCP; M = Mediwork data in FCP; U* = A balanced subset of the UKB validate set.)}\label{ablatable1}
\begin{tabular}{|p{82px}|p{40px}|p{40px}|p{40px}|p{40px}|p{40px}|p{40px}|}
\hline
\multirow{2}{*}{Method}&\multirow{2}{*}{$R^2_{CVD}$}&\multicolumn{2}{c|}{$R^2_{(T,M)}$}&\multicolumn{3}{c|}{MK-MMD (Pre-Ada.)}\\\cline{3-7}
&&Pre-ada.&Post-ada.&T$\leftrightarrow$M&U*$\leftrightarrow$M&U*$\leftrightarrow$T\\ 
\hline
Baseline&0.5586&\textbf{0.3610}&0.4071&\textbf{0.3028}&0.4898&0.1836\\
Weight1&0.5319&0.3263&0.4373&0.3996&0.3789&0.3372\\
Weight2&0.5643&0.1572&0.4176&0.3456&0.3972&0.2333\\
Weight2+SimSiam&0.0267&-0.0128&0&0.0002&4.2678&0.0003\\
Weight2+CLFA&\textbf{0.5703}&0.2144&\textbf{0.4937}&0.3741&\textbf{0.3557}&\textbf{0.1597}\\ %
\hline
\end{tabular}
\end{table}

\begin{figure}[t]
     \centering
     \begin{subfigure}[b]{0.55\textwidth}
         \centering
         \includegraphics[width=\textwidth]{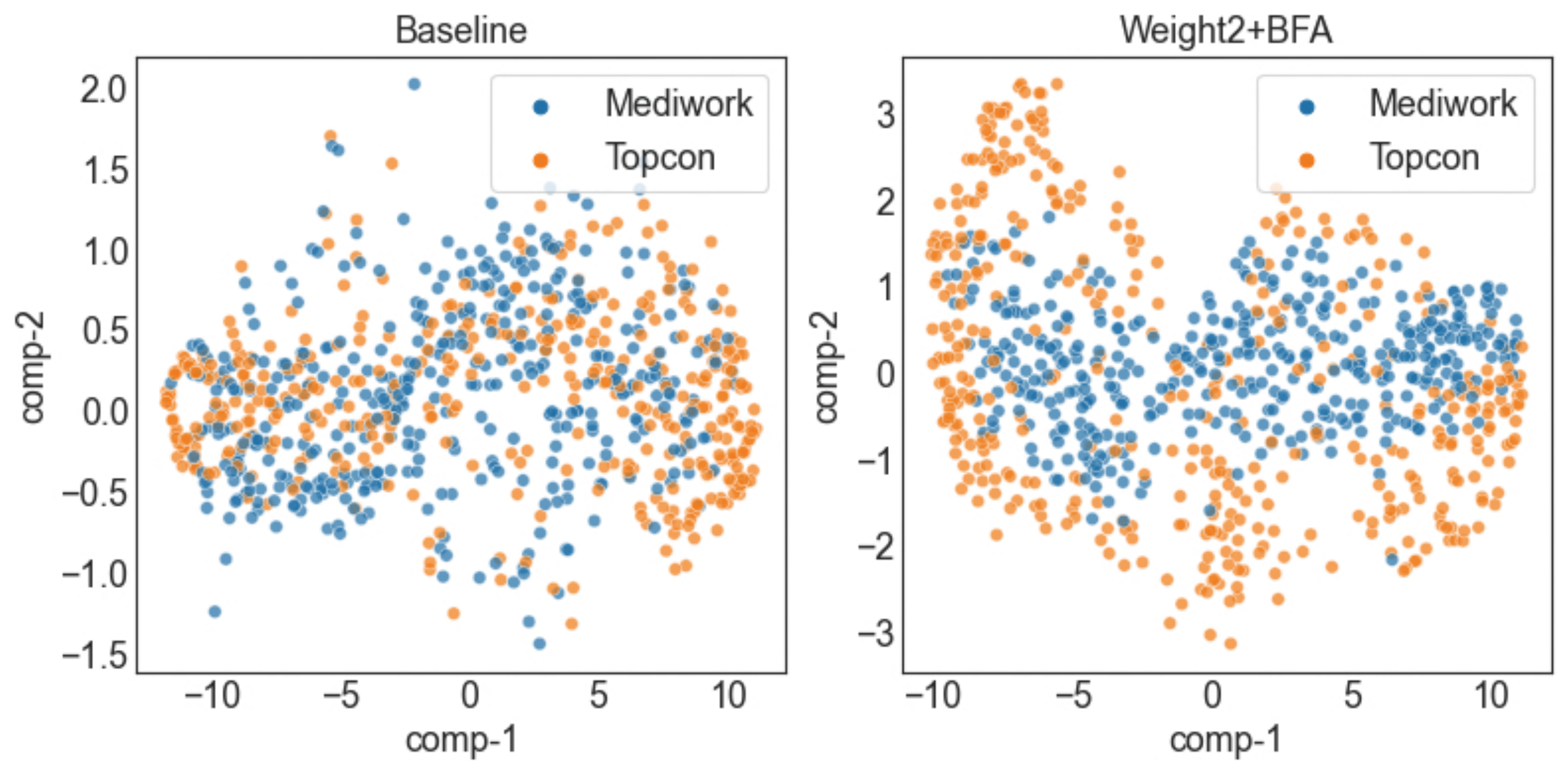}
         \caption{The t-SNE features visualization on the FCP dataset}
         \label{visualization}
     \end{subfigure}
     \hfill
     \begin{subfigure}[b]{0.43\textwidth}
        \begin{tabular}{|c|c|c|}
        \hline
        Model&$AUC_{lat.}$&$AUC_{camera}$\\
        \hline
        Baseline&0.5270&0.6414\\
        Weight2+CLFA&\textbf{0.7322}&\textbf{0.9722}\\
        \hline
        \multicolumn{3}{c}{}\\
        \multicolumn{3}{c}{}\\
        \end{tabular}
        \caption{Laterality and camera prediction performance on the FCP dataset}
        \label{predictca}
    \end{subfigure}
    \caption{Study of features extracted from the pre-training models}
    \label{further}
    %\multicolumn{7}{p{325px}}{\small{U* = A balanced subset of the UKB validate set. The results in this table is without the camera adaptor module.}}
\end{figure}
%\noindent\textbf{Qualitative analysis}
To further investigate the factors for model adaptability, we visualize the image feature of the FCP dataset extracted by ``Baseline'' and ``Weight2+CLFA'' through t-SNE as shown in Fig.~\ref{visualization}. We figure out ``Weight2+CLFA'' has better feature clustering over different cameras. To quantify this observation, we train a single layer fully connected neural network on the image feature to predict the eye laterality and camera. The result is as Tab.~\ref{predictca}, and it proves the ``Weight2+CLFA'' feature can support better discrimination of laterality and camera. Furthermore, it indicates that discrimination on camera may positively affect post-adaptation result consistency.
%The before-adaptation vs. after-adaptation comparison in Tab.\ref{ablatable1} has show our camera adaptor module can improve result coefficient on all tested pre-trained models. We also conduct ablation study of its structure and training loss function based on the ``Weight2+CLFA'' model above.
%We add module design and loss function to find the best combination. To compared the feature-distribution methods, we also try the MK-MMD as loss function.
\begin{table}[t!]
\caption{Ablation study on the camera adaptor (SA=Self-attention.)}\label{ablatable2}
\centering
\begin{tabular}{|p{80px}|p{62px}|p{62px}|p{62px}|p{62px}|}
\hline
\multirow{2}{*}{Module}&\multicolumn{4}{c|}{$R^2(T,{M})$}\\
\cline{2-5}
&$\ell_{CVD}$&$\ell_{feature}$&$\ell_{MK-MMD}$&$\ell_{CVD}+\ell_{feature}$\\ 
\hline
MLP&0.3831&0.3479&0.1349&\textbf{0.3602}\\
SA Block&0.3229&0.3097&\textbf{0.3052}&0\\
SA Block+MLP&\textbf{0.4937}&\textbf{0.4885}&0.2113&0.3458\\
\hline
%\multicolumn{5}{p{320px}}{}
\end{tabular}
\end{table}
%Pre ada, Post ada, TODO
\subsubsection{Camera Adaptor Module}\label{ablation2}
To verify the design of our camera adaptor module, we test it under several different settings with a range of loss functions.  
The results in Tab.~\ref{ablatable2} demonstrate that both the self-attention block and the MLP projector are essential for our camera adaptor module. The comparison of the loss function shows that the WHO-CVD score-focused loss function has the best results. We also observe that optimizing the feature distribution discrepancy (MK-MMD) between features does not improve the prediction result consistency. 
\subsection{Quantitative Analysis}
\begin{table}[t!]
    \centering
    \caption{The overall performance comparison (CVD = WHO-CVD regression on UKB; T=Topcon data in FCP; M = Mediwork data in FCP.)}\label{fcptable}
    \begin{tabular}{|p{210px}|p{60px}|p{60px}|}
    \hline
    Method&$R^2$(CVD)&$R^2(T,{M})$\\
    % \multirow{2}{*}{Method}&\multicolumn{2}{c|}{$R^2$}\\\cline{2-3}
    % &CVD&TC$\rightarrow$MW\\ 
    \hline
    Baseline&0.5586&0.3610\\
    Baseline+DAN~\cite{long2015learning}&0.5586&0.3243\\
    Baseline+DANN~\cite{ajakan2014domain}&-0.0253&0.0254\\
    Baseline+MDD~\cite{zhang2019bridging}&-0.0104&1.000\\
    Baseline+Pix2Pix GAN~\cite{isola2017image}&0.5586&0.3183\\
    Baseline+Cycle GAN~\cite{zhu2017unpaired}&0.5586&0.3594\\
    %Baseline+SimSiamese~\cite{SimSiam}&0.0267&-0.0128\\ 
    Ours (CLFA Pre-training + Adaptation Module)&\textbf{0.5703}&\textbf{0.4732}\\ % cdt2
    %Ours&0.5812&\\ % cdf_t4
    \hline
    %\multicolumn{3}{p{330px}}{\small{CVD = WHO-CVD regression on UKB; T=Topcon data in FCP; M = Mediwork data in FCP.}}
    \end{tabular}
\end{table}
To compare our proposed method and other approaches, we select several typical approaches of feature-alignment and image-alignment. The baseline model in this experiment is as Sec.~\ref{ablation1} and is without the adaptor.
For feature-alignment, we test the DAN~\cite{long2015learning} which optimize the MK-MMD between transformed features, the DANN~\cite{ajakan2014domain} which apply the adversairal training, and the MDD~\cite{zhang2019bridging} which using an auxiliary classifier to optimize the discrepancy between the two domains. For image alignment, we select the Pix2Pix GAN~\cite{isola2017image} and Cycle GAN~\cite{zhu2017unpaired} as the generators. The comparison shows that our method improves the result consistency over the two cameras. However, DAN, DANN, Cycle GAN, and Pix2Pix GAN lead to a lower consistency than baseline. The MDD reach $R^2 = 1.0$ as the model output a collapsed result.
% As there is no ground truth for FCP dataset, we freeze the baseline model and only train the added modules to maintain the source model. 
For the DAN, the reason for degrading may be DAN is designed for unpaired data and its MK-MMD-based loss function have no advantage when pair-wise data is available, which has also been proved in our ablation study (Sec.~\ref{ablatable2}). The DANN and MDD require the parameter updating on the backbone model and lead to the degrading of backbone model. For the Pix2Pix GAN, we check its generated images and find the generator focusing on color toning or image style adjustment with a loss of the microvascular vessels' detail, which is supposed to be an essential visual representation of CVD. The Cycle GAN learns some image style transformation but fails to improve the predcition outcome. The advantage of our CLFA pre-training plus camera adaptor method is that both generalization and adaptation are considered to improve the model's adaptability. 
%\ZY{The results analysis are horrible, please improve immediately,

\section{Conclusion}
This paper researches the domain discrepancy problem in developing a fundus-image-based CVD risk predicting algorithm. We observe that the deep learning model trained conventionally will have a variant representation on photos from different fundus cameras. Therefore, we propose a cross-laterality feature learning training method and a camera adaptor module. The experiments show that our design has improved on the prediction result consistency. Also, we find that the feature-space distribution discrepancy between pre-training and target domain data may be the key factor of model transportability. Future research will explore the data augmentation in adaption to overcome the data lacking and utilize the FCP data in the backbone model pre-training.  

\bibliographystyle{splncs04}
\bibliography{paper1145}

\end{document}